\documentclass[letterpaper]{article} % DO NOT CHANGE THIS
\usepackage{aaai2026}  % DO NOT CHANGE THIS
\usepackage{times}  % DO NOT CHANGE THIS
\usepackage{helvet}  % DO NOT CHANGE THIS
\usepackage{courier}  % DO NOT CHANGE THIS
\usepackage[hyphens]{url}  % DO NOT CHANGE THIS
\usepackage{graphicx} % DO NOT CHANGE THIS
\usepackage{verbatim}

\usepackage{multirow}
\usepackage{booktabs}  % if you’re using \toprule, \midrule, etc.
\usepackage{tcolorbox}
\usepackage{float}
\usepackage{soul,xcolor}      % 加载 soul 用于下划线，xcolor 用于颜色
\setul{0.5ex}{0.3ex}          % （可选）调整下划线深度和厚度
\newcommand{\colorul}[2][red]{%
  \setulcolor{#1}%            % 设置下划线颜色为第一个可选参数
  \ul{#2}%                    % 对第二个参数的文字加下划线
}

\usepackage{natbib}  % DO NOT CHANGE THIS AND DO NOT ADD ANY OPTIONS TO IT
\usepackage{caption} % DO NOT CHANGE THIS AND DO NOT ADD ANY OPTIONS TO IT
\usepackage{amsmath, amssymb, mathtools}

\usepackage{amsmath}   % 核心数学公式支持
\usepackage{amssymb}   % 提供更多数学符号, 如 \mathbb
\usepackage{amsfonts}  % 提供数学字体

\usepackage{booktabs}  % 用于高质量表格的 \toprule, \midrule, \bottomrule
\usepackage{multirow}  % 用于表格中的多行合并
\usepackage[table]{xcolor}   % 用于给表格单元格和行上色
\usepackage{amsmath}
\usepackage{adjustbox}
\usepackage{adjustbox}
\usepackage{enumitem}

\usepackage[utf8]{inputenc} % allow utf-8 input
\usepackage[T1]{fontenc}    % use 8-bit T1 fonts
\usepackage{url}            % simple URL typesetting
\usepackage{booktabs}       % professional-quality tables
\usepackage{amsfonts}       % blackboard math symbols
\usepackage{nicefrac}       % compact symbols for 1/2, etc.
\usepackage{microtype}      % microtypography
\usepackage{xcolor}         % colors
\usepackage{graphicx}
\usepackage{amsmath}
\usepackage[table]{xcolor} 

\usepackage{algorithm}

\usepackage{newfloat}
\usepackage{listings}
\usepackage[noend]{algpseudocode}

\DeclareCaptionStyle{ruled}{labelfont=normalfont,labelsep=colon,strut=off} % DO NOT CHANGE THIS
\floatstyle{ruled}
\newfloat{listing}{tb}{lst}{}
\floatname{listing}{Listing}

\title{LLM-CAS: Dynamic Neuron Perturbation for Real-Time Hallucination Correction}
\author{
Jusheng Zhang$^{1}$,
Ningyuan Liu$^{1}$,
Yijia Fan$^{1}$,
Zihao Huang$^{1}$, \\
Qinglin Zeng$^{1}$,
Kaitong Cai$^{1}$,
Jian Wang$^{2}$,
Keze Wang$^{1}$\thanks{Corresponding author.} \\[3pt]
$^{1}$ Sun Yat-sen University.
$^{2}$ Snap Inc.
}

\usepackage{bibentry}
\begin{document}

\maketitle

\begin{abstract}
Large language models (LLMs) often generate hallucinated content lacking factual or contextual grounding, hindering their reliability in critical applications. Traditional methods like supervised fine-tuning and reinforcement learning from human feedback are data-intensive and computationally expensive, while static parameter editing struggles with context-dependent errors and catastrophic forgetting. To overcome these limitations, we introduce LLM-CAS, a framework that formulates real-time hallucination correction as a hierarchical reinforcement learning (HRL) problem. LLM-CAS trains an agent to learn a sophisticated policy, dynamically selecting optimal, temporary neuron perturbations during inference based on the immediate context. This learned, policy-driven approach provides greater adaptability than prior dynamic methods that rely on heuristic or pre-defined adjustments. As a result, LLM-CAS achieves significant performance gains across various LLMs, improving accuracy by 10.98 percentage points on StoryCloze, 2.71 points on TriviaQA, and 2.06 points on TruthfulQA's MC1 score, thereby outperforming static methods like ITI and CAA, as well as the dynamic SADI framework. This context-aware, efficient approach promises enhanced reliability for LLMs in high-stakes domains, with future potential for multimodal extensions.
\end{abstract}

\section{Introduction}
\noindent Large Language Models (LLMs) \cite{touvron2023llama2openfoundation,radford2019language,DBLP:journals/corr/abs-2005-14165,Z1,Z2,A1,m1} represent a transformative force in technology, demonstrating remarkable capabilities in natural language understanding and generation. However, their full potential is curtailed by a pervasive and critical flaw: ``hallucination'' \cite{hall2,mckenna2023sourceshallucinationlargelanguage,Z3,Z4,Z5,A2,m2}. This tendency to generate content that is factually incorrect or contextually ungrounded \citep{Huang_2025} remains a formidable obstacle to their reliable deployment in mission-critical applications. While traditional mitigation strategies such as Supervised Fine-Tuning (SFT) \citep{fan2024preferenceorientedsupervisedfinetuningfavoring,Z6,Z7,A3,A5,m3} and Reinforcement Learning from Human Feedback (RLHF) \citep{ouyang2022traininglanguagemodelsfollow,casper2023openproblemsfundamentallimitations,Z8,Z9,A4,m4} have shown some efficacy, they are often hampered by a reliance on large-scale, high-quality annotated data \cite{DBLP:journals/corr/abs-2106-09685,Z10,Z11,m5}. Furthermore, these methods can suffer from diminished generalization or inadvertently introduce new biases, and the prohibitive computational cost of full-model fine-tuning renders them impractical for many scenarios.

\medskip
\noindent To circumvent the high costs of retraining, a significant body of research has explored more granular interventions, such as directly modifying internal parameters or activation states to rectify specific knowledge deficits. Many of these approaches follow a ``locate-then-edit'' paradigm \citep{meng2023locatingeditingfactualassociations, DBLP:journals/corr/abs-2104-08696,A6,m6}. They first identify the model parameters $W$ most relevant to a target fact, often via causal tracing, and then compute and apply a one-off, static perturbation $\Delta W$. The objective is to force the edited model $M(x; W_{\mathrm{edited}})$ to produce an updated output $V_1$ for a given input $K_1$, while preserving its original behavior on unrelated inputs $K_0$. This is often formalized as:
$W_{\mathrm{edited}} = W + \Delta$
$\Delta = \arg\min_{\tilde{\Delta}} \Bigl(\bigl\|(W + \tilde{\Delta}) K_1 - V_1\bigr\|^2 + \lambda \bigl\|\tilde{\Delta} K_0\bigr\|^2\Bigr)$
However, despite their utility for isolated corrections, such static edits prove brittle when faced with widespread, context-dependent hallucinations. Even effective methods inevitably introduce perturbations that negatively impact unrelated knowledge \citep{meng2023masseditingmemorytransformer,A7,m7}. These deleterious effects accumulate with sequential edits, risking catastrophic forgetting or model collapse.

\medskip
\noindent While static edits have clear limitations, recent work has shifted towards dynamic interventions that occur during inference, avoiding permanent parameter changes. However, these approaches often rely on heuristic or pre-defined adjustments, which can lack the adaptability needed for complex, context-dependent hallucinations. To address this gap, we introduce a more principled and adaptive framework. Specifically, we are the first to frame the challenge of real-time correction as a \textit{hierarchical reinforcement learning} (HRL) problem \citep{DBLP:journals/corr/KulkarniNST16, 10.1023/A:1025696116075}. Our central hypothesis is that temporary, context-specific perturbations can effectively correct errant outputs without inflicting permanent damage on the model's integrity. For a given input $x$ that elicits a hallucinated output $y_h = M(x; W)$, instead of seeking a universal parameter update, we train a policy $\pi(a \mid s)$ to dynamically generate an optimal, context-specific perturbation $\Delta_{\mathrm{dyn}}$. The policy's action $a$ is conditioned on a state $s$ that encodes the input $x$, the hallucinated output $y_h$, and a small set of reference examples $S_{\mathrm{small}}$. This action guides the generation of $\Delta_{\mathrm{dyn}}$, which is temporarily applied to the model's neuron activations, yielding a corrected output:
$y_c = M\bigl(x; W \oplus \Delta_{\mathrm{dyn}}\bigr)$
where $\oplus$ denotes the perturbation operation. The HRL structure allows the agent to make structured, multi-level decisions, efficiently exploring the vast perturbation space to learn fine-grained correction capabilities. The agent's goal is to maximize a reward $R_t$ tied to the factual accuracy and quality of $y_c$. Our formal objective is to learn a hierarchical policy $\pi(a \mid s)$ that can autonomously apply the most effective $\Delta_{\mathrm{dyn}}$ for any given hallucination scenario, such that:
$M\bigl(x; W \oplus \Delta_{\mathrm{dyn}}\bigr)\;\longrightarrow\;y_{\mathrm{correct}}$
By harnessing this dynamic learning mechanism, our approach flexibly addresses diverse hallucination types while maintaining model generality and minimizing unintended side effects. In this paper, we present and evaluate this Hierarchical Reinforcement Learning-based Dynamic Neuron Perturbation framework. We demonstrate that by training an agent to make context-aware intervention decisions online, adjusting both scope and intensity in real-time. We will open-source our method to support the robust deployment of LLMs in critical applications.
\begin{figure*}[h]
% \vskip 0.2in
\centering
\includegraphics[width=\linewidth]{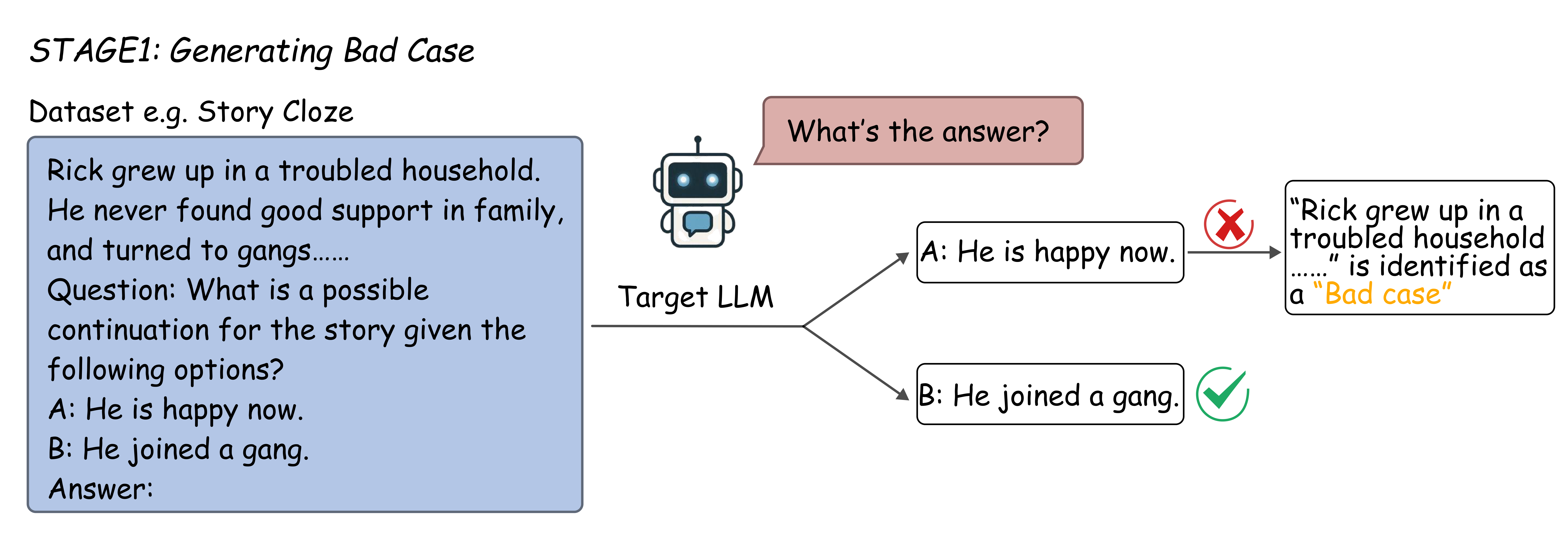} % Assuming Frame_842.pdf is in the same directory or a specified path
\caption{Stage 1: A Story Cloze example \cite{mostafazadeh-etal-2016-corpus}, i.e., prefix “Rick grew up in a troubled household\ldots”, question, and endings A: “He is happy now.” and B: “He joined a gang.” (correct), fed to the target LLM.}
\label{stage1}
\end{figure*}
\section{Related Works}

% REVISED:
% I've slightly rephrased this to be more encompassing and to create a smoother
% transition to why newer methods are needed. The core message and citations are preserved.
\paragraph{Strategies for Mitigating LLM Hallucinations.}
\noindent Foundational approaches to curb hallucinations in large language models (LLMs) primarily involve large-scale training or fine-tuning. These include supervised fine-tuning (SFT) on high-quality, factual data \cite{radford2019language,dettmers2023qloraefficientfinetuningquantized,kang2024unfamiliarfinetuningexamplescontrol,Z12} and reinforcement learning from human feedback (RLHF), which aligns model behavior with human preferences \cite{christiano2023deepreinforcementlearninghuman}. While effective to an extent, SFT often struggles with generalization to out-of-domain facts, and RLHF is notoriously data-intensive, requiring extensive human annotation and labor \cite{DBLP:journals/corr/abs-2112-00861}. A prominent challenge for both is the risk of catastrophic forgetting or performance degradation in general capabilities. To circumvent these issues, more targeted intervention methods have been developed.

% REVISED:
% Renamed to ``Model Editing'' for broader scope. I've sharpened the distinction
% between editing *permanent weights* versus our method of editing *temporary activations*.
\paragraph{Model Editing for Factual Correction.}
\noindent Model editing techniques aim to directly modify an LLM's parameters to inject or correct specific factual knowledge. A dominant paradigm is ``locate-then-edit,'' where methods first identify the neurons or parameters most relevant to a piece of knowledge and then apply a calculated, one-off update \cite{DBLP:journals/corr/abs-2110-11309, meng2023locatingeditingfactualassociations, meng2023masseditingmemorytransformer}. These approaches apply *static and permanent* perturbations to the model's weights. Consequently, they often struggle with context-dependent hallucinations and risk accumulating negative side effects that degrade unrelated knowledge, especially when edits are applied sequentially \cite{hase-etal-2023-methods}. In sharp contrast, LLM-CAS avoids permanent parameter modification altogether, instead performing *dynamic and temporary* perturbations on neuron activations during inference, offering a more adaptive and less disruptive solution.

% NEW:
% This is the most CRITICAL addition. It directly addresses the review feedback
% by situating your work among other dynamic methods and explicitly
% distinguishing it from SADI.
\paragraph{Dynamic Inference-Time Interventions.}
\noindent More recent efforts have pivoted towards dynamic interventions that occur only during inference, leaving the base model's weights untouched. Methods like Inference-Time Intervention (ITI) \cite{li2024inferencetimeinterventionelicitingtruthful} and Contrastive Activation Addition (CAA) \cite{panickssery2024steeringllama2contrastive} steer model behavior by adding a fixed, pre-computed steering vector to activations at specific layers. While dynamic, these vectors are typically static across different inputs. A closer related work is SADI \cite{wang2025semanticsadaptiveactivationinterventionllms}, which proposes using \textbf{semantics-adaptive steering vectors} that can change based on the input. However, SADI's mechanism for generating these vectors often relies on pre-defined rules or simpler optimization. Our work builds upon this trajectory but introduces a key distinction: we formulate the problem within a formal \textbf{Hierarchical Reinforcement Learning (HRL)} framework \cite{10.1023/A:1025696116075, DBLP:journals/corr/KulkarniNST16}. LLM-CAS does not use a pre-defined steering mechanism but instead *learns* a sophisticated, multi-level policy $\pi(a|s)$ to select the optimal intervention strategy in real-time, offering a more principled and powerful approach to adaptation.

% REVISED:
% Re-titled and refocused to emphasize the HIERARCHICAL aspect, which is a
% key part of your contribution. This makes the paragraph less about PPO in
% general and more about *why you use it in a hierarchical setup*.
\paragraph{Hierarchical Policy Optimization with PPO.}
\noindent To navigate the vast and complex action space of neuron perturbations, we employ a hierarchical learning structure optimized with Proximal Policy Optimization (PPO) \cite{DBLP:journals/corr/SchulmanWDRK17}. PPO is a policy gradient algorithm known for its stability and sample efficiency, making it well-suited for complex control tasks compared to other RL algorithms \cite{black2024trainingdiffusionmodelsreinforcement,mnih2015humanlevel,DBLP:journals/corr/SchulmanLMJA15,DBLP:journals/corr/MnihBMGLHSK16}. The novelty of our approach lies in the hierarchical application: a high-level policy selects a macro-level intervention target, while a low-level policy determines the fine-grained perturbation details. 
\section{Dynamic Neuron Perturbation}

\noindent To counteract the pervasive issue of hallucination in large language models (LLMs), where generated content may be factually inconsistent or lack contextual support, we introduce a novel framework centered on dynamic neuron perturbation. This framework leverages hierarchical reinforcement learning (HRL) to train an agent that learns to apply optimal, context-aware perturbations to specific neuron activations during LLM inference. By doing so, it enables the online correction of potential hallucinations in real-time. This section details the formal problem definition, the architectural design of our framework, the mechanics of its key components, and the underlying learning algorithms.

\subsection{Problem Definition}
\label{sec:reward_definition}

\noindent Given a pretrained large language model $M$ with parameters $W$, an input sequence $x$ may elicit a hallucinated output $y_h = M(x; W)$, as depicted in the ``Bad cases'' of Figure~\ref{stage1}. Our objective is to learn a hierarchical policy $\pi(a|s)$ that, conditioned on the current state $s$ (which includes $x$ and model history), dynamically selects an optimal perturbation action $a$. This action, in turn, guides the generation of a temporary, context-specific perturbation vector $\Delta_{dyn}$. This vector is then precisely applied to the activation states of a targeted set of neurons, denoted $Act(x; W)$. The post-intervention model then produces a corrected, high-quality output $y_c$, where the perturbation is formally applied as $Act_{perturbed} \leftarrow Act(x;W) \oplus \Delta_{dyn}$ (with $\oplus$ representing the perturbation operation). The goal is for $y_c$ to minimize hallucination while preserving semantic coherence, relevance, and fluency.

To achieve this, we formulate the hallucination correction task as a Markov Decision Process (MDP). The agent's goal is to learn an optimal hierarchical policy $\pi^*$ that maximizes the expected cumulative discounted reward:
\begin{equation}
\label{eq:reward_formula}
\pi^* = \arg\max_{\pi} \mathbb{E}_{\tau \sim \pi} \left[ \sum_{t=0}^{T} \gamma^t R_{t+1} \right],
\end{equation}
where $\tau = (s_0, a_0, R_1, s_1, a_1, \dots)$ is a trajectory or episode, $\gamma \in [0,1]$ is the discount factor, and $R_{t+1}$ is the reward received after executing action $a_t$ and transitioning to state $s_{t+1}$. For a rigorous definition of the state, action, and transition components, see Appendix A.

Our dynamic neuron perturbation method, illustrated in Figure \ref{stage2}, forms the core solution to this MDP. It integrates five key components:
\begin{itemize}
    \item \textbf{Target LLM ($M$):} The model to be corrected, which provides baseline outputs and responds to the applied perturbations.
    \item \textbf{Hierarchical RL Agent:} Comprised of high- and low-level PPO agents, responsible for learning and executing the hierarchical policy $\pi(a|s)$.
    \item \textbf{Dynamic Neuron Perturbation Environment:} An interface that constructs the state $s_t$, translates the agent's action $a_t$ into a concrete perturbation, and computes the resulting reward $R_{t+1}$.
    \item \textbf{Adaptive Perturbation Localization Module:} The mechanism that translates the abstract action $a_t$ into the specific numerical perturbation $\Delta_{dyn}$, utilizing macro-function networks, a learnable dynamic mask, and neuron attribution analysis \cite{DBLP:journals/corr/SundararajanTY17, DBLP:journals/corr/abs-1711-06104}.
    \item \textbf{LLM Response Evaluation Module:} A component that assesses the quality of the corrected output $y_c$ to generate the reward signal $R_{t+1}$, which in turn drives the learning of both the RL agent and the adaptive mask.
\end{itemize}
During training, this system operates in a closed loop: for each ``bad case'' $x$, the agent observes the state, selects an action to guide a perturbation, receives a reward based on the corrected output, and updates its policy. This dynamic, learning-based approach stands in contrast to static model editing methods \cite{yao2023editinglargelanguagemodels}, which compute a single, permanent parameter update $\Delta W^*$ based on a fixed optimization objective:
\begin{multline}
\Delta W^* = \arg\min_{\tilde{\Delta W}}
\Bigl(
  \mathcal{L}_{edit}\bigl((W+\tilde{\Delta W})K_1, V_1\bigr) \\
  + \lambda\,\mathcal{L}_{preserve}(\tilde{\Delta W}K_0)
\Bigr),
\end{multline}
where $K_1, V_1$ represent target knowledge and $K_0$ represents knowledge to be preserved. The inherent static nature of these methods makes them struggle with context-dependent hallucinations, thereby highlighting the necessity of our dynamic approach.

\subsection{Dynamic Neuron Perturbation Environment}
\label{sec:environment}
The agent interacts with the LLM through a purpose-built environment, which provides a state representation $s_t \in S$ at each timestep $t$. The state vector $s_t$ is a concatenation of four components: \textbf{Input Context Embedding ($Emb(x)$)}, which encodes the semantic features of the input; \textbf{Baseline Model Performance ($Scores_{baseline}$)}, a set of metrics (e.g., hallucination, relevance, fluency) for the LLM's unperturbed output $y_h$; \textbf{Current Best Performance ($Scores_{best}$)}, which tracks the highest-quality scores achieved so far within the episode; and a \textbf{Normalized Step Count ($Steps_{norm}$)}, which indicates the progress of the interaction. Thus, $s_t = \text{concat}(Emb(x), Scores_{baseline}, Scores_{best}, Steps_{norm})$.

From state $s_t$, the agent selects an action $a_t \in A$ from a hierarchical discrete action space. This decouples the decision into two levels:
\begin{itemize}
    \item \textbf{High-Level Action ($a_H \in A_H$):} Selects a macro-level target, specifically a network category $C_k$ from a predefined set $A_H = \{C_1, C_2, \dots, C_{N_H}\}$ that corresponds to functional clusters of neurons.
    \item \textbf{Low-Level Action ($a_L = (a_L^{type}, a_L^{mag})$):} Given the high-level choice $a_H$, this action specifies the fine-grained intervention details: the \textbf{Perturbation Type} ($a_L^{type} \in \{\text{noise, zero, scale, \dots}\}$) and the \textbf{Perturbation Magnitude} ($a_L^{mag} \in \{m_1, m_2, \dots, m_{N_M}\}$).
\end{itemize}
The complete action is the tuple $a_t = (a_H, a_L^{type}, a_L^{mag})$. Upon execution of $a_t$, the environment facilitates the state transition $P(s_{t+1}|s_t,a_t)$. The perturbation $\Delta_{dyn}$ defined by $a_t$ is applied to the LLM's activations, leading to a new output $y_c$. This output is evaluated to yield current scores, $Scores_{current}$. The environment then updates $Scores_{best}$ and increments $Steps_{norm}$ to form the next state $s_{t+1}$. This transition is near-deterministic, with stochasticity arising primarily from the LLM's decoding process.

The environment provides a scalar reward $R_t = R(s_t, a_t, s_{t+1})$ calculated as:
\begin{equation}
R_t = w_h \cdot \Delta Score_{h,t} + w_r \cdot \Delta Score_{r,t} + w_f \cdot \Delta Score_{f,t} + R_{exp,t}
\end{equation}
Here, $\Delta Score_{h,t} = Score_{h,baseline} - Score_{h,current,t}$ quantifies the reduction in hallucination. Concurrently, $\Delta Score_{r,t}$ and $\Delta Score_{f,t}$ measure the change in relevance and fluency, respectively. The weights $w_h, w_r, w_f$ balance these competing objectives. An exploration bonus, $R_{exp,t} > 0$, is added to incentivize discovering new strategies, particularly when the current action fails to improve upon the best-known hallucination score.

\subsection{Hierarchical Reinforcement Learning Agent}
\label{sec:hrl_agent}

We employ a hierarchical reinforcement learning (HRL) framework powered by Proximal Policy Optimization (PPO), an algorithm selected for its sample efficiency and robust training stability in complex decision-making domains. The agent's architecture is bifurcated into two tiers, i.e., high-level and low-level, each implemented with its own actor (policy) and critic (value) networks to effectively manage macro and micro decisions.

The \textbf{high-level component} governs strategic, macro-level choices. Its policy network, $\pi_H(a_H|s; \theta_{\pi_H})$, is an MLP that maps the state $s$ to a probability distribution over the macro target categories $a_H \in A_H$. The corresponding value network, $V_H(s; \theta_{V_H})$, also an MLP, estimates the expected cumulative return from state $s$.
\begin{equation}
\scalebox{0.8}{$
\begin{split}
\pi_H(A_H \mid s; \theta_{\pi_H})
      &= \underbrace{\operatorname{Softmax}\!
         \bigl(
           \operatorname{MLP}_{\pi_H}(s;\, \theta_{\pi_H})
         \bigr)}_{\text{High-Level Policy Network}}
         \\[2pt]
V_H(s; \theta_{V_H})
      &= \underbrace{\operatorname{MLP}_{V_H}\!
         \bigl(
           s;\, \theta_{V_H}
         \bigr)}_{\text{High-Level Value Network}}
\end{split}
$}
\label{eq:high_level_policy_value}
\end{equation}
Theoretical details of our hierarchical PPO are in Appendix A. and hyperparameter settings are in Appendix H. The \textbf{low-level component} makes tactical, micro-level decisions under the guidance of the high-level action. Its policy network, $\pi_L(a_L|s, a_H; \theta_{\pi_L})$, takes both the state $s$ and the chosen high-level action $a_H$ (via its embedding) as input. It then outputs a probability distribution over the specific perturbation types and magnitudes $a_L \in A_L$. The low-level value network, $V_L(s, a_H; \theta_{V_L})$, estimates the expected return for being in state $s$ having committed to macro action $a_H$.
\begin{equation}
\scalebox{0.7}{$
\begin{split}
\pi_L\!\bigl(A_L \mid s, a_H; \theta_{\pi_L}\bigr)
      &= \underbrace{\operatorname{Softmax}\!
         \Bigl(
           \operatorname{MLP}_{\pi_L}\!
           \bigl(
             \operatorname{concat}(s,\;\operatorname{embed}(a_H));
             \theta_{\pi_L}
           \bigr)
         \Bigr)}_{\text{Low-Level Policy Network}}
         \\[2pt]
V_L\!\bigl(s, a_H; \theta_{V_L}\bigr)
      &= \underbrace{\operatorname{MLP}_{V_L}\!
         \Bigl(
           \operatorname{concat}(s,\;\operatorname{embed}(a_H));
           \theta_{V_L}
         \Bigr)}_{\text{Low-Level Value Network}}
\end{split}
$}
\label{eq:low_level_policy_value}
\end{equation}
The \textbf{learning process} consists of two phases: experience collection and network updates. In each episode, the agent executes the combined action $a_t = (a_{H,t}, a_{L,t})$ determined by its policies, receives a reward $R_t$, and transitions to the next state $s_{t+1}$. The resulting experience tuple, $(s_t, a_{H,t}, a_{L,t}, R_t, s_{t+1}, \log \pi_H(a_{H,t}|s_t), \log \pi_L(a_{L,t}|s_t, a_{H,t}))$, is stored in separate high- and low-level replay buffers. During the update phase, both policy layers are optimized using PPO's clipped surrogate objective:
\begin{equation}
\small
L^{\text{CLIP}}(\theta_{\pi}) = -\mathbb{E}_t \left[ \min\left( r_t(\theta_{\pi})\hat{A}_t, \text{clip}(r_t(\theta_{\pi}), 1-\epsilon, 1+\epsilon)\hat{A}_t \right) \right]
\end{equation}
where $r_t(\theta_{\pi}) = \frac{\pi_{\theta_{\pi}}(a_t|s_t)}{\pi_{\theta_{\pi,\text{old}}}(a_t|s_t)}$ is the importance sampling ratio. The advantage $\hat{A}_t$ is estimated using Generalized Advantage Estimation (GAE) \cite{schulman2018highdimensionalcontinuouscontrolusing}. The total loss function also includes a squared-error value loss $L^{\text{VF}}(\theta_V)$ and a policy entropy term $S[\pi_{\theta_{\pi}}(s_t)]$ to encourage exploration and prevent premature policy convergence.

\begin{figure*}[t]
% \vskip 0.2in
\centering
\includegraphics[width=0.8\linewidth]{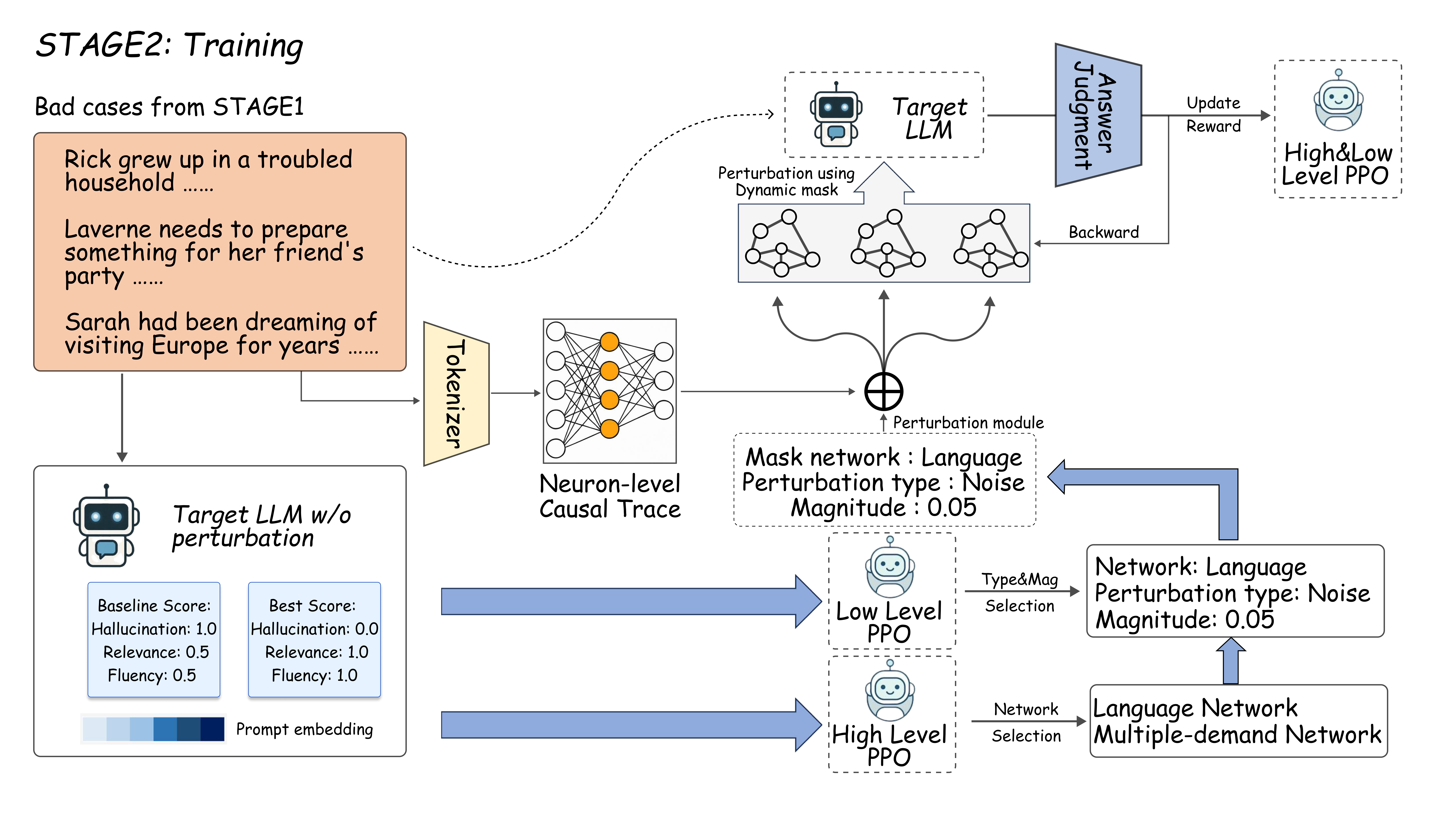} % Assuming Frame_843.pdf is available
\vspace{-20pt}
\caption{Stage 2: Training ``bad cases from Stage 1 undergo neuron‐level causal tracing to generate perturbation masks on the LLM’s representations; the perturbed inputs are re‐evaluated to produce a reward for optimizing two PPO agents.}
\label{stage2}
% \vskip -0.2in
\end{figure*}

\subsection{Adaptive Masking and Neuron-level Causal Trace}
\label{sec:adaptive_masking_causal_trace}

\noindent A core challenge in our framework is translating the agent's abstract action into a precise and minimally invasive neuron-level intervention. We address this through a two-stage adaptive masking mechanism, which integrates insights from neuron-level causal tracing (see Figure~\ref{stage2}) and shares mask definitions with prior work like llm-localization~\cite{alkhamissi2025llmlanguagenetworkneuroscientific}.

The process begins when the high-level policy selects a macro-functional network category $C_k$ (e.g., $C_{lang}$ for a ``Language Network''). This provides a semantically meaningful, high-level target for intervention. The core of our mechanism then unfolds in two stages:

\textbf{1. Learning a General Sparse Mask.} For each category $C_k$ and layer $l$, we introduce a \textbf{learnable dynamic mask}, $M_{k,l}$, parameterized by $\theta_{k,l}$. This mask learns a general, input-agnostic pattern of which neurons are most often relevant for correction within that functional block. It applies a gating function to produce a selection strength $M_{k,l}(i) \in [0,1]$ for each neuron $i$:
$M_{k,l}(i; \theta_{k,l}, \tau_{gate}) = \sigma(\theta_{k,l}(i) / \tau_{gate})$
The mask parameters for the entire model, $\theta_{mask} = \{\theta_{k,l}\}$, are trained via an independent optimizer to minimize the loss function $\mathcal{L}_{mask}$:
\begin{equation}
\scalebox{0.8}{$
\begin{split}
\mathcal{L}_{mask}(\theta_{mask})
  &= \underbrace{-\mathbb{E}_{\text{episode}}\bigl[R_{total\_ep}\bigr]}_{\text{Negative Expected Reward}}
    \\[2pt]
  &\quad
    +\,\underbrace{\lambda_{sparse}\sum_{k,l}\|M_{k,l}\|_{1}}_{\text{L1 Sparsity Penalty}}
    \;+\;
    \underbrace{\lambda_{L0}\sum_{k,l}\bigl\|\mathbb{I}(M_{k,l}>\epsilon_{th})\bigr\|_{0}}_{\text{L0 Sparsity Penalty}}
\end{split}
$}
\label{eq:mask_loss}
\end{equation}
This objective forces the mask to be both effective (by maximizing the total episode reward $R_{total\_ep}$) and sparse. Sparsity is enforced by the L1 and approximate L0 regularization terms, which penalize the magnitude and number of active neurons, respectively, thereby minimizing potential interference.

\textbf{2. Input-Specific Adaptation.} To tailor the intervention to the current input, a \textbf{Neuron-level Causal Trace module} computes neuron attribution scores, $Attr_l(x) \in \mathbb{R}^{d_l}$, for each layer $l$ using methods like Integrated Gradients. These scores represent the ``critical activation patterns'' specific to input $x$. The final \textbf{operational mask}, $M_{op,k,l}$, is produced by dynamically modulating the general mask with these real-time attribution scores:
\begin{equation}
M_{op, k,l}(i) = M_{k,l}(i; \theta_{k,l}) \odot \text{normalize}(|\text{Attr}_l(x, i)|)
\end{equation}

where $\text{normalize}(\cdot)$ scales the absolute attribution values to a $[0, 1]$ range and $\odot$ denotes element-wise multiplication. This two-stage approach, learning a general sparse template and then adapting it with input-specific causal information, enables targeted, real-time perturbations without the need to retrain $\theta_{mask}$ for every new input.

\subsection{Model Output Evaluation and Feedback Mechanism}
\label{sec:output_evaluation}

\noindent The efficacy of the RL agent is critically dependent on a high-quality feedback signal. This is provided by the model output evaluation module (referenced in Figure~\ref{stage1} as ''Answer Judgment''), which assesses the target LLM's corrected output $y_c$ along three dimensions: hallucination (H), relevance (R), and fluency (F). The resulting numerical scores, $\{Score_H, Score_R, Score_F\}$, are essential inputs for calculating the agent's reward $R_t$ and for optimizing the dynamic mask parameters $\theta_{\text{mask}}$.

For the nuanced demands of open-ended generation tasks, this evaluation primarily employs \textbf{Llama2-7B-Instruct} \cite{touvron2023llama2openfoundation} as a ''judging LLM.'' This choice aligns with the recent and growing trend of using capable LLMs as scalable evaluators \cite{zheng2023judgingllmasajudgemtbenchchatbot}. We direct the judge using a meticulously crafted prompt, $\text{prompt}_{\text{eval}}(x, y_c)$, which instructs it to score the output $y_c$ given the original input $x$:
\begin{equation}
(\text{Score}_H, \text{Score}_R, \text{Score}_F) = \text{LLM}(\text{prompt}_{\text{eval}}(x, y_c))
\end{equation}
For multiple-choice tasks, the reward signal is derived more directly and objectively from task-specific metrics, such as the correctness of the selected option, as detailed in our experimental setup.

We acknowledge and proactively address the potential for inherent biases or errors in any LLM-based judge \cite{wang2023largelanguagemodelsfair}. To mitigate this risk, the evaluation prompt, $\text{prompt}_{\text{eval}}$, is iteratively refined to improve its objectivity. Furthermore, by designing the reward function to depend on score \textit{changes} (e.g., $\Delta Score_{H,t}$) rather than absolute values, we reduce the impact of any systemic scoring bias from the judge. While a full-scale human-alignment study is beyond the scope of this work, our initial qualitative checks revealed a reasonable correlation between the judge's scores and human assessments for the error types we target. This feedback mechanism is thus structured to guide the coordinated optimization of the entire framework towards an effective and robust hallucination mitigation strategy.
\section{Experiments}
\textbf{Experimental Setup:}
\label{sec:experimental_setup}
We evaluate our LLM-CAS on both Multiple-choice and Open-ended generation tasks, using a comprehensive set of datasets in each case to ensure the generalizability of LLM-CAS. To better compare our LLM-CAS with SADI \cite{wang2025semanticsadaptiveactivationinterventionllms}, we use the same evaluation method as SADI. All experiments are conducted on eight NVIDIA A100 GPUs. 

\textbf{Multiple-choice Tasks} For the Multiple-choice tasks, we use the Story Cloze \cite{DBLP:journals/corr/MostafazadehCHP16}, SST-2 \cite{socher-etal-2013-recursive}, BoolQ \cite{clark-etal-2019-boolq}, and Winogrande \cite{DBLP:journals/corr/abs-1907-10641} datasets. These datasets feature between 2 and 5 answer choices, primarily focused on distinguishing ``correct'' from ``incorrect'' options. We format each question with the correct answer as a prompt, then extract the logits from the LLM’s response to determine its predicted choice, which is used for scoring. 

\textbf{Open-Ended Generation Tasks.} For open-ended generation tasks, we employ the TriviaQA \cite{joshi-etal-2017-triviaqa}, ToxiGen \cite{hartvigsen-etal-2022-toxigen}, and TruthfulQA \cite{DBLP:journals/corr/abs-2109-07958} datasets. Detailed descriptions and data splits for each dataset are available in Appendix C. Additionally, we include the multiple-choice variant of TruthfulQA to evaluate the MC-Score. For TriviaQA, we use Exact Match as the evaluation metric to assess the capabilities of the LLM-CAS framework. For ToxiGen and TruthfulQA, we use fine-tuned LLMs to evaluate the factual correctness of generated outputs. Specifically, we use \texttt{toxigen\_hatebert} (based on HateBERT \cite{caselli-etal-2021-hatebert} and ToxiGen data) for ToxiGen; for TruthfulQA, we use \texttt{truthfulqa-truth-judge-llama2-7B} to evaluate factual correctness, and \texttt{truthfulqa-info-judge-llama2-7B} to evaluate informativeness (these judges are based on LLaMA2 \cite{touvron2023llama2openfoundation} and the TruthfulQA dataset). All these models can be found on Hugging Face. They have been deployed since they are fine-tuned for judging, alleviating their hallucinations. % Corrected from \newline to \\ for consistency, or prefer paragraph breaks

\textbf{Target LLMs} Our primary baseline model is LLaMA2-7B-CHAT \cite{touvron2023llama2openfoundation}. To assess the generalizability of the LLM-CAS framework, we conduct experiments across LLMs with different architectures and parameter scales. For architectural diversity, we test MISTRAL-7B \cite{jiang2023mistral7b} and Gemma-1.1-7b-it \cite{gemmateam2024gemmaopenmodelsbased}. 

\textbf{Baseline Comparisons} To better illustrate the effectiveness of our LLM-CAS, we compare it against Inference-Time Intervention (ITI) \cite{li2024inferencetimeinterventionelicitingtruthful}, Contrastive Activation Addition (CAA) \cite{panickssery2024steeringllama2contrastive}, and SADI \cite{wang2025semanticsadaptiveactivationinterventionllms}. 
\vspace{-1em}
\subsection{Main results}
\subsubsection{Multiple-choice Questions}

\begin{table}[ht]
  \centering
  \scriptsize
  \setlength{\tabcolsep}{3pt}            % 缩小列间距
  \caption{Accuracy of \textsc{llm-CAS} on multiple‐choice tasks.}
  \label{tab:mc_results}
  % 用 tabular* 指定宽度为当前栏宽（\linewidth），并让列间距自动伸缩
  \begin{tabular*}{\linewidth}{@{\extracolsep{\fill}} lccccc}
    \toprule
    Task   & StoryCloze & SST-2  & BoolQ  & Winogrande & Average \\
    \midrule
    Baseline & 65.06   & 88.63 & 70.52 & 50.91 & 68.78 \\
    ITI      & 68.50   & \textbf{91.38} & 74.10 & 52.80 & 71.70 \\
    CAA      & 74.65   & 91.16 & \textbf{74.98} & 52.64 & 73.36 \\
    SADI     & 67.57   & 88.69 & 70.40 & 51.93 & 69.65 \\
    \textbf{Ours} & \textbf{76.04} & 91.30 & 74.47 & \textbf{52.90} & \textbf{73.68} \\
    \bottomrule
  \end{tabular*}
\end{table}

\textbf{Effectiveness on Multiple-Choice Tasks} As shown in Table~\ref{tab:mc_results}, \textsc{llm-CAS} consistently outperforms the baseline model and all other competing methods across a variety of multiple-choice question datasets, demonstrating its effectiveness in improving the accuracy of discrete choice tasks. Unlike ITI and CAA, which apply static, vector-based perturbations to neurons, both SADI and \textsc{llm-CAS} employ dynamic perturbation strategies. Moreover, the superior accuracy of \textsc{llm-CAS} validates the correctness of its dynamic masking combined with a PPO-based optimization, which is a clear advantage over SADI's dynamic interventions. Notably, \textsc{llm-CAS} achieves a 10.98\% absolute improvement over the baseline on the Story Cloze dataset, underscoring its strong potential.

\subsubsection{Open-ended Generation Questions}
\begin{table}[ht]
  \centering
  \scriptsize
  \setlength{\tabcolsep}{2pt} % reduce column separation
  \caption{Performance of \textsc{llm-CAS} on open‐ended generation tasks. TruthfulQA’s MC scores use the multiple‐choice format; True and Info use open‐ended evaluation.}
  \label{tab:open-ended}
  \begin{tabular}{lcc cccccc}
    \toprule
    \multirow{2}{*}{Task}
      & TriviaQA       & ToxiGen            & \multicolumn{6}{c}{TruthfulQA} \\
      \cmidrule(lr){2-2} \cmidrule(lr){3-3} \cmidrule(lr){4-9}
      & EM             & toxicity$\downarrow$ & True & Info & True$\times$Info & MC1  & MC2  & MC3  \\
    \midrule
    Baseline & 41.60 & 49.71 & 66.83 & \textbf{99.51} & 66.50 & 33.41 & 51.07 & 24.76 \\
    ITI      & 42.80 & 45.27 &    –    &    –    &    –    & 34.64 & 51.55 & 25.32 \\
    CAA      & 43.20 & 49.71 & 71.60 & 83.84 & 60.03 & 34.03 & \textbf{52.76} & 25.62 \\
    SADI     & 43.50 & \textbf{17.14} & 74.54 & 93.51 & 69.71 & 34.88 & 52.50 & 25.79 \\
    Ours     & \textbf{44.31} & 47.63 & \textbf{75.12} & 94.22 & \textbf{70.78} & \textbf{35.47} & 51.45 & \textbf{26.43} \\
    \bottomrule
  \end{tabular}
\end{table}

\textbf{\textsc{llm-CAS} improves the performance on open-ended generation tasks.} \textsc{llm-CAS} remarkably improves performance on open-ended generation tasks. To further assess its effectiveness, we evaluate \textsc{llm-CAS} on several benchmark datasets. As illustrated in Table~\ref{tab:open-ended}, compared to the Baseline models, the \textsc{llm-CAS} framework enhances the performance of LLMs across multiple dimensions, including knowledge-intensive QA (TriviaQA), safety/toxicity control (ToxiGen), and truthfulness (TruthfulQA). Notably, \textsc{llm-CAS} improves the accuracy on TriviaQA by 2.71 points (from 41.60 to 44.31), and reduces toxicity in ToxiGen by 2.08 points (toxicity score from 49.71 to 47.63). Furthermore, it significantly boosts performance on TruthfulQA, increasing the MC1 score from 33.41 to 35.47 (+2.06) and the MC2 score from 51.07 to 51.45 (+0.38). These improvements suggest that \textsc{llm-CAS} not only enhances factual correctness and safety but also outperforms existing steering-based methods on certain datasets. This effectiveness can be attributed to the integration of dynamic masking and the Proximal Policy Optimization (PPO) algorithm within \textsc{llm-CAS}, which endows it with strong exploration capabilities. 
\begin{figure}[h]
% \vskip 0.2in
\begin{center}
\centerline{\includegraphics[width=\linewidth]{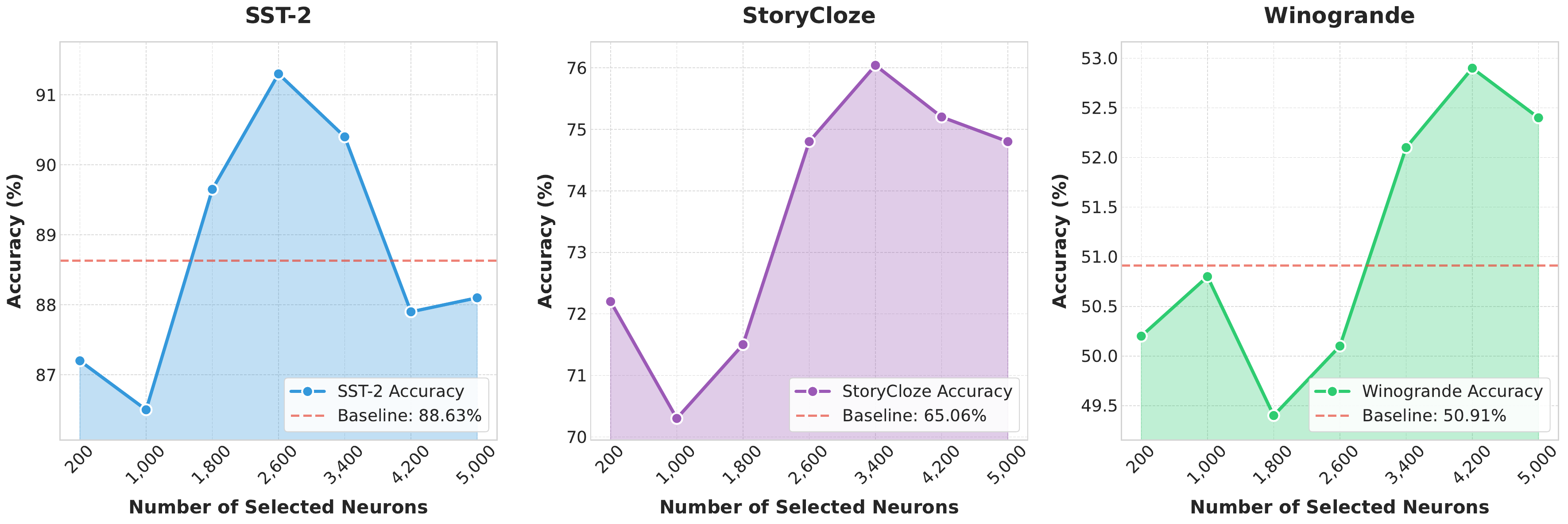}}
\caption{Impact of perturbing different numbers of selected neurons on the outcomes. The Adaptive Mask dynamically adjusts both the number and positions of neurons to identify the optimal perturbation strategy.}
\label{selected_neurons}
\end{center}
% \vskip -0.2in
\end{figure}

\begin{figure}[htbp]
% \vskip 0.2in
\centering
\includegraphics[width=\linewidth]{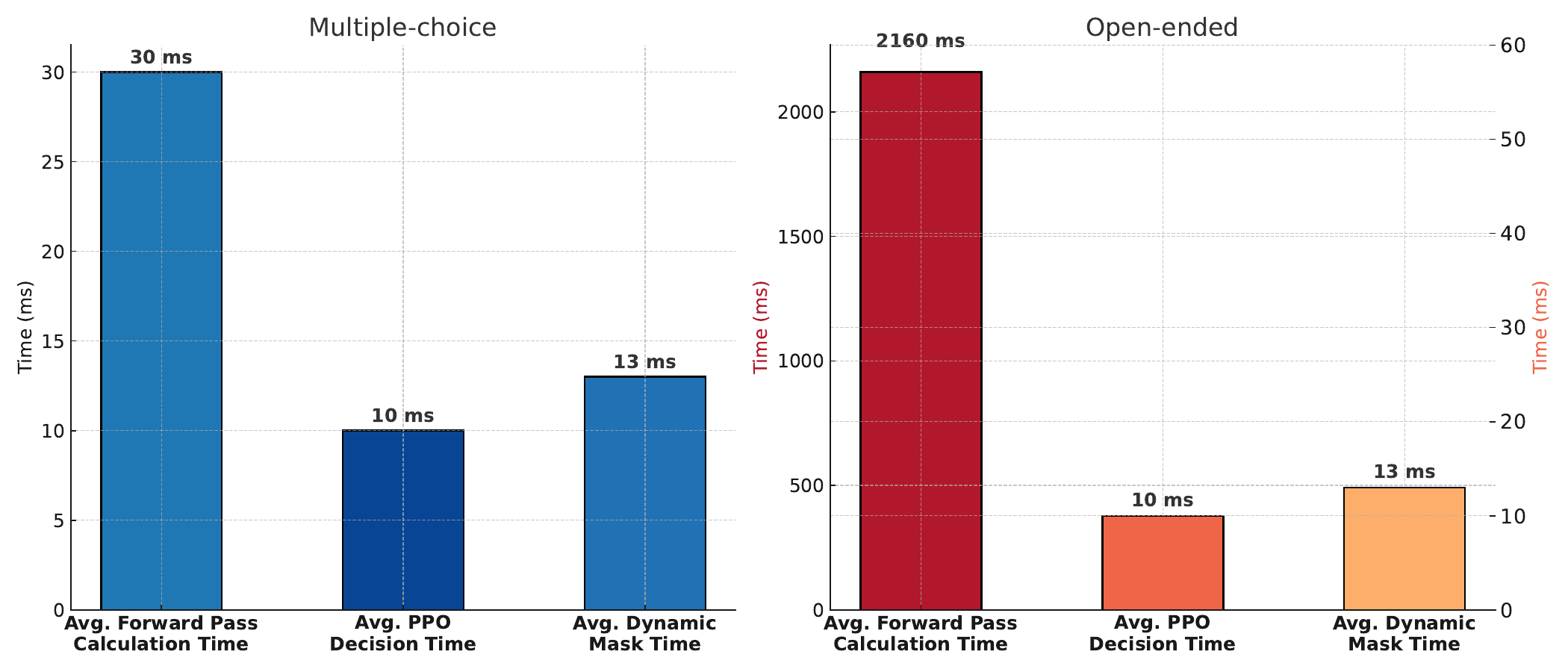}
\caption{Comparison of PPO decision time, dynamic mask inference time, and overall model execution time. For multiple-choice tasks, ``forward pass time'' denotes the duration of the model’s forward propagation to obtain logits; for open-ended tasks, ``generate time'' denotes the duration of calling \texttt{AutoModelForCausalLM.generate}\cite{wolf-etal-2020-transformers} for text generation without sampling.}
\label{time_comparison}
% \vskip -0.2in
\end{figure}
\begin{figure}[h]
% \vskip 0.2in
\vspace{-1em}
\centering
\includegraphics[width=\linewidth]{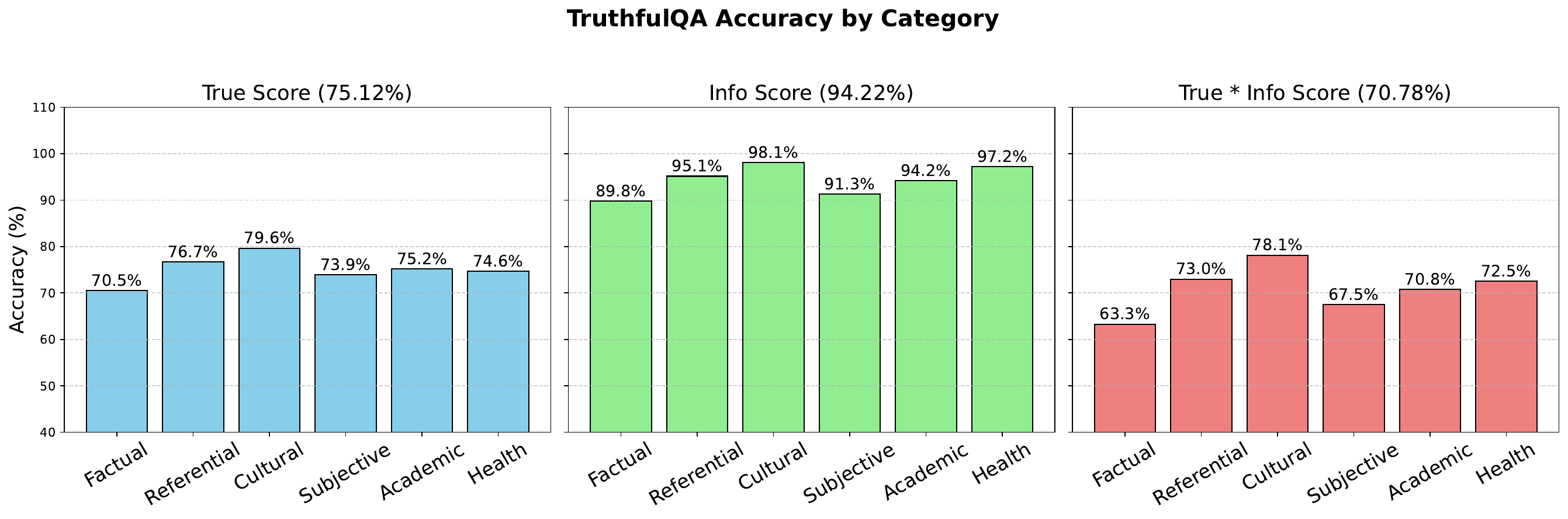}
\caption{Accuracy across six TruthfulQA hallucination categories. \textsc{llm-CAS} delivers substantial gains in Cultural accuracy, while improvements in Factual are modest.}
\label{tqa_performance}
% \vskip -0.2in
\end{figure}
\vspace{-1em}
\subsection{Evaluation of LLM-CAS on Various Language Models}
Although \textsc{llm-CAS} achieves strong results on the widely used Llama2-7B-Chat model, its effectiveness on other LLM architectures remains unclear. To evaluate the generalizability of the \textsc{llm-CAS} framework across different model families, we conduct experiments on two open-source, Transformer-based LLMs, i.e., Mistral-7B-Instruct-v0.3 and Gemma-1.1-7b-it. The results, shown in Table~\ref{tab:arch_results}, indicate that on both Mistral and Gemma, \textsc{llm-CAS} consistently improves accuracy over the baseline on the StoryCloze, SST-2, and Winogrande datasets, demonstrating its robustness to architectural variation. Notably, the largest gains are observed on StoryCloze for both models, suggesting that the neuron-level interventions provided by \textsc{llm-CAS} are particularly effective for tasks closely tied to narrative coherence.
\begin{table}[htbp]
  \centering
  \caption{Performance of different models on various tasks with and without LLM-CAS}
  \label{tab:arch_results}
  % 把宽度设为列宽的 80%，高度保持比例
  \resizebox{1\columnwidth}{!}{%
    \begin{tabular}{lccc}
      \toprule
      \textbf{Model}                  & \textbf{StoryCloze} & \textbf{SST-2} & \textbf{Winogrande} \\
      \midrule
      Mistral-7B-Instruct-v0.3        & 21.51               & 90.33          & 58.96               \\
      Gemma-1.1-7b-it                 & 60.95               & 74.08          & 48.46               \\
      Mistral-7B w/ LLM-CAS           & 34.41               & 90.45          & 59.80               \\
      Gemma-1.1 w/ LLM-CAS            & 69.76               & 79.19          & 49.71               \\
      \bottomrule
    \end{tabular}%
  }
\end{table}

\subsection{Ablation Studies}

\begin{table}[htbp]
\centering
\caption{Ablation results of \textsc{llm-CAS} on multiple-choice tasks. Each row removes a key component.}
\label{tab:ablation_new}
\resizebox{1\columnwidth}{!}{%
\begin{tabular}{lcccc|c}
\toprule
\textbf{Variant} & \textbf{SST-2} & \textbf{BoolQ} & \textbf{Winogrande} & \textbf{StoryCloze} & \textbf{Average} \\
\midrule
\textsc{llm-CAS} (Full)              & \textbf{91.30} & \textbf{74.47} & \textbf{52.90} & \textbf{76.04} & \textbf{73.68} \\
Random mask                          & 86.73          & 67.10          & 51.32          & 70.20          & 68.84          \\
Random action                        & 82.45          & 64.32          & 49.15          & 66.87          & 65.70          \\
Random mask and action               & 80.18          & 62.05          & 47.98          & 63.41          & 63.41          \\
\bottomrule
\end{tabular}
}
\end{table}

\textbf{Dynamic Masking is Critical.} Removing the dynamic masking mechanism and replacing it with random neuron selection leads to a noticeable drop in performance across multiple tasks, e.g., a decrease of 5.84 points on StoryCloze and 7.37 points on BoolQ. This demonstrates that adaptively identifying task‐relevant neurons during inference is crucial for accurate decision‐making.
\textbf{PPO Optimization Enhances Adaptability.} Replacing PPO with non‐adaptive optimization (i.e., random action selection) results in a substantial performance decline, e.g., –9.17 points on StoryCloze and –10.15 points on BoolQ. This confirms that PPO’s policy, gradient approach better navigates the complex activation space of transformer‐based models, enabling more effective control.
\textbf{Combined Effect is Greater Than the Sum.} When both dynamic masking and PPO are removed, the model suffers the largest degradation, with the average accuracy dropping to 63.41 (–10.27 from full). This shows that the two components reinforce each other: masking enables targeted neural modulation, while PPO learns robust control policies for adaptation.
These findings indicate that both dynamic masking and PPO‐based optimization are essential to the effectiveness of \textsc{llm-CAS}. Their combination achieves significantly better performance across diverse classification tasks compared to using either in isolation. Further ablation studies on reward functions and other design choices are included in Appendix D.
\section{Conclusion}
This paper presents LLM-CAS, a dynamic neuron-perturbation framework that uses hierarchical reinforcement learning to apply temporary, context-aware tweaks during inference, i.e., correcting hallucinations in real-time without harming general model behavior. By combining adaptive masking with neuron-level causal tracing, it precisely targets only the activations that cause errors. Framed as an RL problem balancing factuality, relevance, and fluency, our LLM-CAS outperforms both static edits and other dynamic schemes across multiple classification and generation benchmarks. Ablation studies confirm that its dynamic masking and PPO-based policy are both essential for robust correction.

\section{Acknowledgments}
This work was supported in part by the National Natural Science Foundation of China (NSFC) under Grant 62276283, in part by the China Meteorological Administration's Science and Technology Project under Grant CMAJBGS202517, in part by Guangdong Basic and Applied Basic Research Foundation under Grant 2023A1515012985, in part by Guangdong-Hong Kong-Macao Greater Bay Area Meteorological Technology Collaborative Research Project under Grant GHMA2024Z04, in part by Fundamental Research Funds for the Central Universities, Sun Yat-sen University under Grant 23hytd006, and in part by Guangdong Provincial High-Level Young Talent Program under Grant RL2024-151-2-11.
\bibliography{aaai2026}

\end{document}